\documentclass[runningheads]{llncs}

\usepackage{eccv}

\usepackage{eccvabbrv}

\usepackage{graphicx}
\usepackage{booktabs}
\usepackage{xspace}
\usepackage{multirow}
\usepackage{makecell}
\usepackage{array}
\usepackage{algorithm}
\usepackage{algpseudocode}
\usepackage{balance}
\usepackage{comment}
\usepackage{amsmath, amssymb, amsfonts}
\newtheorem{prop}{Proposition}
\usepackage{booktabs}
\usepackage{graphicx}
\usepackage[accsupp]{axessibility}
\usepackage[most]{tcolorbox}

\usepackage{hyperref}

\usepackage{orcidlink}

\begin{document}

\title{Dual-Margin Embedding for Fine-Grained Long-Tailed Plant Taxonomy}

\author{
Cheng-Yaw Low\inst{1,2}$^{*}$
\and
Heejoon Koo\inst{2}$^{*}$
\and
Jaewoo Park\inst{3}
\and
Meeyoung Cha\inst{2}
}

\institute{
Changwon National University, Republic of Korea
\and
Max Planck Institute for Security and Privacy (MPI-SP), Germany
\and
AiV Co., Republic of Korea
}

\maketitle

\begingroup
\renewcommand{\thefootnote}{\fnsymbol{footnote}}
\footnotetext[1]{This work was done when the first and second authors were with MPI-SP.}
\endgroup

\begin{abstract}
Taxonomic classification of ecological families, genera, and species underpins biodiversity monitoring and conservation. Existing computer vision methods typically address fine-grained recognition and long-tailed learning in isolation. However, additional challenges such as spatiotemporal domain shift, hierarchical taxonomic structure, and previously unseen taxa often co-occur in real-world deployment, leading to brittle performance under open-world conditions. We propose TaxoNet, an embedding learning framework with a theoretically grounded dual-margin objective that reshapes class decision boundaries under class imbalance to improve fine-grained discrimination while strengthening rare-class representation geometry. We evaluate TaxoNet in open-world settings that capture co-occurring recognition challenges. Leveraging diverse plant datasets, including Google Auto-Arborist (urban tree imagery), iNaturalist (Plantae observations across heterogeneous ecosystems), and NAFlora-Mini (herbarium collections), we demonstrate that TaxoNet consistently outperforms strong baselines, including multimodal foundation models.
\end{abstract}
    
\vspace{-5mm}
\section{Introduction}
\label{sec:intro}
Biodiversity forms the foundation for the stability and resilience of ecosystems, ensuring essential ecological functions, regulating the climate, and supporting human well-being \cite{sala2000global}. 
Its conservation is recognized by the United Nations as fundamental, offering both a scientific rationale and a policy framework for global action. This commitment is reflected in several United Nations Sustainable Development Goals (SDGs); SDG 15 `Life on Land' which directly targets biodiversity loss, alongside SDG 13 `Climate Action', and SDG 11 `Sustainable Cities and Communities' \cite{scbd2017biodiversity}. In parallel, machine learning and computer vision have become transformative tools for biodiversity monitoring, particularly for ecological taxonomic classification \cite{duraiappah2011intergovernmental, waldchen2018automated}.
However, practical deployment remains challenging due to environmental heterogeneity and data scarcity. These complexities necessitate robust systems capable of generalizing to real-world ecological conditions.

\begin{figure}[t]
    \centering
    \begin{minipage}{0.7\linewidth}
        \centering
        \includegraphics[width=\linewidth]{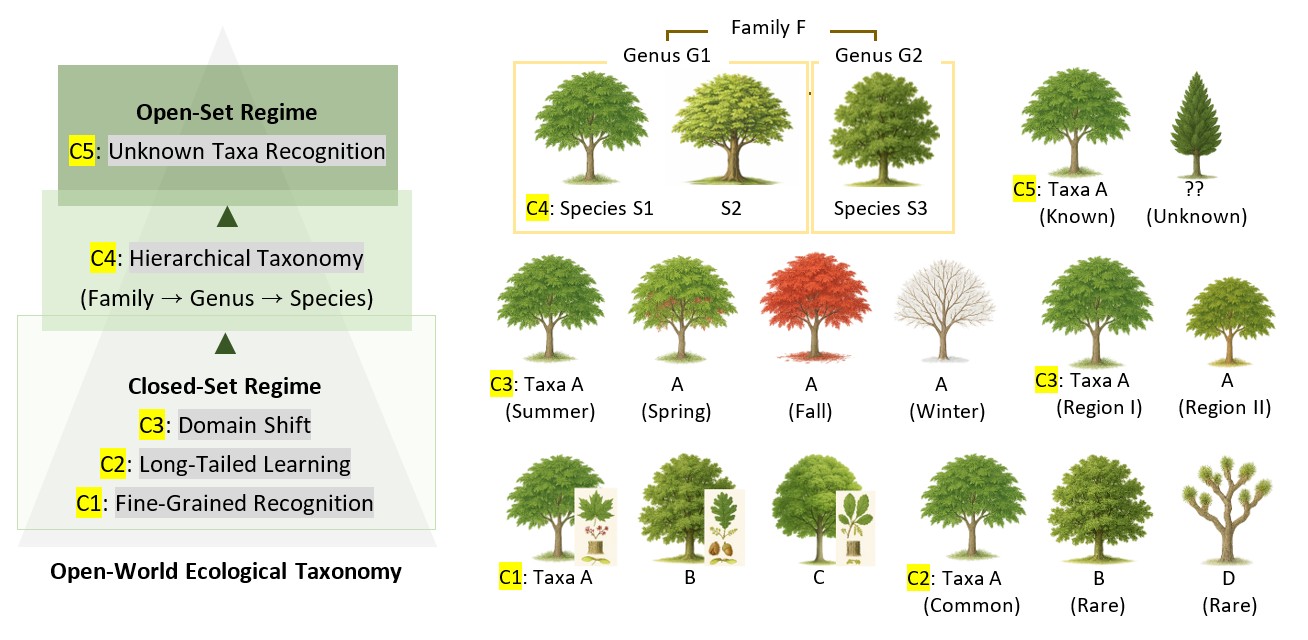}
    \end{minipage}
    \hfill
    \begin{minipage}{0.28\linewidth}
    \caption{Prior work in ecological taxonomy studies fine-grained and long-tailed recognition under closed-set assumptions, whereas real-world settings involve spatiotemporal shifts, hierarchical taxonomic structure, and unseen taxa.}
    \label{fig:openworld}
    \end{minipage}
\end{figure}

Real-world ecological data often exhibit long-tailed distributions, where only a few common (\ie head) taxa dominate observations, while rare (\ie tail) taxa—many of which are endemic, endangered, or invasive—remain underrepresented \cite{van2018inaturalist, beery2022auto, park2024naflora}. This bias causes recognition models to favor dominant taxa, limiting performance on those most critical for conservation and management. Moreover, taxonomic classification requires fine-grained recognition, as many taxa differ only by subtle morphological traits. For example, species within the same genus (\eg \emph{Acer rubrum} vs. \emph{Acer saccharum}) may exhibit high inter-taxa similarity and intra-taxa variability, leading to misclassification~\cite{van2018inaturalist}.

Beyond long-tailed class distribution, ecological data exhibit test-time shifts driven by seasonal, habitat, and geographic variability. These factors introduce mismatches between training and deployment conditions that degrade generalization \cite{beery2022auto}. Biodiversity monitoring is fundamentally an open-set problem due to undocumented taxa in the wild, yet most models rely on closed-set assumptions \cite{low2025open}. This results in the misclassification and poses risks to conservation efforts \cite{park2021divergent}. While existing research tackles long-tail imbalance via resampling, loss reweighting \cite{johnson2019survey, lin2017focal, cui2019class}, and more recently, logit adjustments \cite{menon2020long} or margin-based classifiers \cite{cao2019learning}, these  targeted solutions limit generalizability in realistic conditions, where class imbalance, fine-grained variability, open-set dynamics, and distribution shifts intersect.


We study ecological taxonomy under realistic deployment conditions, where fine-grained recognition, long-tailed class distributions, spatiotemporal variation, and open-set scenarios co-occur. Departing from the standard practice of examining these challenges independently, we analyze model robustness in-the-wild where these factors jointly manifest (see Figure~\ref{fig:openworld}). Although ecological taxonomy spans diverse groups, including insects, birds, mammals, and fungi commonly represented on large-scale biodiversity platforms such as iNaturalist \cite{van2018inaturalist}, our work focuses on plant-level taxa across urban, natural, and specimen-based environments. This focus is motivated by the availability of large-scale annotated datasets and the foundational role of plants in biodiversity monitoring and conservation efforts \cite{blackmore2024botanic}.

Our primary contributions are as follows:
\vspace{-3mm}
\begin{itemize}
\item We propose a theoretically grounded dual-margin loss that models class imbalance for fine-grained and long-tailed recognition, forming the foundation of our embedding learning framework, TaxoNet. Specifically, the dual-margin formulation suppresses repulsive gradients on tail-class prototypes induced by dominant head classes, leading to more balanced decision boundaries.

\item We introduce a novel sample selection strategy that prioritizes tail taxa and morphologically diverse instances reflecting broader ecological variation.

\item Evaluation shows advances in fine-grained recognition under long-tailed distributions, domain shift, and hierarchical structure against strong baselines and multimodal foundation models. 

\end{itemize}

\noindent Our contributions support ecological AI applications for robust biodiversity recognition,  Henceforth, the term \emph{taxa} refers to taxonomic ranks including family, genus, and species, unless otherwise specified. 

\section{Related Work}
We review prior work along five axes: fine-grained recognition, long-tailed learning, domain generalization, hierarchical taxonomy, and open-set recognition. While fine-grained and long-tailed recognition have been extensively studied in isolation under closed-set assumptions, their joint manifestation in plant taxonomic recognition remains underexplored.

\vspace{-2mm}
\paragraph{Fine-Grained Classification.} 
Fine-grained ecological taxonomy requires distinguishing visually similar species based on subtle morphological cues (\eg, petal venation, leaf shape, or bark texture in plants), often relying on high-resolution imagery to preserve discriminative details \cite{van2018inaturalist,cui2018large}. Recent advances have extended this task to the vision–language paradigm, where visual and textual prompts jointly guide classification \cite{sun2023fine,lewis2023gist,stevens2024bioclip}. Although margin-based softmax losses—originally developed for face recognition—offer strong class separation properties \cite{wang2018cosface,deng2019arcface}, they remain underexplored in ecological classification, largely due to instability under long-tailed distributions and limited inductive capacity for rare taxa. These challenges are exacerbated in ecological datasets, where inter-species similarity and intra-species variation are both high.

\vspace{-1mm}
\paragraph{Long-Tailed Learning.}
Ecological datasets, including plant taxa, suffer from severe class imbalance where a few common species dominate and many rare and underrepresented taxa. Data-level strategies, such as oversampling rare taxa or undersampling common taxa, are conventional but risk overfitting or discarding informative data~\cite{johnson2019survey}. Loss reweighting offers an alternative by scaling gradients based on class priors, though naïve implementations can cause training instability~\cite{zhang2023deep}. More principled approaches include Focal Loss \cite{lin2017focal} for  hard examples; Class-Balanced Loss \cite{cui2019class} based on  effective samples size; Label-Distribution-Aware Margin (LDAM) loss  \cite{cao2019learning} for larger  margins on rare classes; and Logit Adjustment \cite{menon2020long} to recalibrate decision boundaries. Despite their efficacy against imbalance, these methods do not address specific ecological challenges like fine-grained discrimination, distributional shift, or taxonomic novelty.

\vspace{-2mm}
\paragraph{Domain Generalization.} 
Ecological deployment involves distributional shifts across geographic regions and temporal conditions. Despite its practical importance, domain generalization has received limited attention in biodiversity recognition. The Google Auto-Arborist dataset \cite{beery2022auto} is one of the few large-scale, street-level tree repositories curated to evaluate inter-city generalization in North America. In contrast, iNaturalist-2019 \cite{van2018inaturalist}, while taxonomically diverse through citizen science contributions, lacks structured test splits for assessing geographic or temporal robustness. Hence, evaluating model robustness under distributional shift remains an open challenge in ecological AI.

\vspace{-2mm}
\paragraph{Hierarchical Taxonomy.}
Ecological labels are inherently structured by biological taxonomy (family–genus–species), forming nested semantic relationships. Leveraging hierarchical metadata, biodiversity datasets such as iNaturalist-2019 \cite{van2018inaturalist} and BIOSCAN-1M \cite{gharaee2023bioscan}, recent foundation models such as BioCLIP \cite{stevens2024bioclip} demonstrate improved taxonomic separability across multiple ranks. However, most benchmarks focus solely on genus or species levels. Consequently, explicit evaluation of hierarchical robustness in ecological recognition remains limited.

\vspace{-2mm}
\paragraph{Open-Set Recognition.}
Most ecological models operate under a closed-set assumption, recognizing only taxa seen during training. This is evident in apps like \emph{Seek Camera} by iNaturalist, \emph{Pl@ntNet}, and \emph{Flora Incognita}, which support roughly 30,000 plant species \cite{waldchen2018automated}. However, real-world biodiversity monitoring often violates this assumption, as hundreds of thousands of vascular plant species have been documented globally \cite{barajas2022global,govaerts2021world}. This gap hinders initiatives like the Kunming-Montreal Global Biodiversity Framework, which mandates comprehensive monitoring by 2030. Addressing this challenge requires open-set recognition frameworks that can detect taxonomic novelty \cite{walter2013openset,park2021divergent}. Although recent studies have explored open-set challenges in ecological domains such as marine and bird species \cite{fishopenset2023,opensetbioacoustics2024}, its application to large-scale plant taxonomy remains limited.
\section{Methodology}
\label{sec:methodology}
We propose TaxoNet, a convolutional embedding framework with a dual-margin loss that reshapes class decision boundaries to improve generalization.

\vspace{-2mm}
\subsection{Preliminary}
\paragraph{Problem Definition.}
Given a $c$-class plant taxonomic classification problem, let \( \mathcal{D}=\{(x_i, y_i)\}_{i=1}^N \) denote a training dataset of $N$ examples, where $x_i$ is the $i$-th input image and $y_i \in \{1, \ldots, c\}$ is the corresponding class label (family, genus, or species). Let \( \phi: \mathbb{R}^{H \times W \times C} \rightarrow \mathbb{R}^d \) be a feature encoder that maps each image \( x_i \) to a $d$-dimensional embedding vector \( \mathbf{x}_i = \phi(x_i) \). For notational simplicity, we omit the sample index \( i \) in subsequent derivations unless otherwise specified.

The additive margin-based softmax cross-entropy loss is formulated as: 
\begin{equation}
\mathcal{L} =
 - \log \frac{e^{s (z_{y} - m)}}
{e^{s (z_{y} - m)} + \sum_{j \ne y} e^{s z_{j}}},
\label{eq:ce}
\end{equation}
where \( s \) is a scaling factor and \( m \) is an additive margin term. 
Each embedding vector \( \mathbf{x} \) is compared with a set of class prototypes \( \{\mathbf{w}_j \in \mathbb{R}^d\}_{j=1}^c \), with logits defined as \( z_j = \mathbf{w}_j^\top \mathbf{x} \), where \( \mathbf{w}_y \) denotes the prototype corresponding to the ground-truth class \( y \).

Setting \( s = 1 \) and \( m = 0 \) reduces Eq.~\eqref{eq:ce} to the standard softmax cross-entropy loss. While effective for general classification tasks, it lacks explicit mechanisms to enforce intra-class compactness and inter-class separation—both essential for distinguishing fine-grained plant taxa with subtle morphological traits such as leaf shape, texture, or color. Moreover, it does not account for the long-tailed class distributions prevalent in plant and other ecological datasets~\cite{van2018inaturalist, beery2022auto}.

An alternative formulation introduces additive margin-based softmax classifiers (AM-Softmax)~\cite{wang2017normface, wang2018cosface}, which incorporate \( s>1 \) and \( m>0 \). Specifically, logits are defined as cosine similarities between normalized embeddings and class prototypes: \( z_j = \cos(\theta_j) = \hat{\mathbf{w}}_j^\top \hat{\mathbf{x}} \), given \( \|\hat{\mathbf{x}}\| = \|\hat{\mathbf{w}}_j\| = 1 \). Although AM-Softmax was not originally designed for long-tailed classification, it has been shown to outperform standard softmax variants in fine-grained recognition tasks such as biometrics~\cite{wang2018cosface,deng2019arcface}, making it a compelling foundation for ecological taxonomy.

\begin{figure}[ht!]
    \centering
    \includegraphics[width=0.8\textwidth]{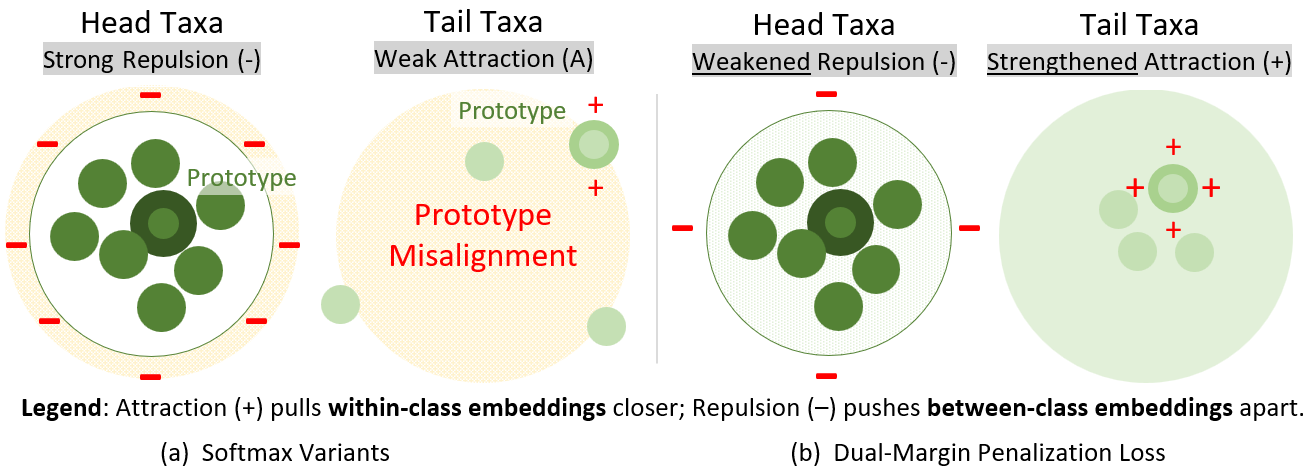}
    \caption{Schematic comparison of softmax-based losses and the proposed dual-margin penalization loss: (a) Softmax losses cause \emph{prototype misalignment} as head classes strongly repel tail embeddings; (b) the proposed dual-margin loss regulates intra-class attraction and inter-class repulsion, inducing compact tail-class embeddings.}    
    \label{fig:embedding_collapse}
\end{figure}

\vspace{-2mm}
\paragraph{Motivation.} 
Softmax-based classifiers like AM-Softmax variants treat all classes with equal importance, implicitly assuming balanced label distributions. Under long-tails, tail-class examples are rarely observed, resulting in gradients from head classes to dominate the training dynamics. This imbalance induces strong attractive (pull) forces that draw head-class embeddings toward their prototypes, while tail-class prototypes experience cumulative repulsive forces from dominant classes. As a result, tail-class prototypes may become persistently misaligned with their corresponding embeddings, a phenomenon we refer to as \emph{prototype misalignment}. This misalignment disrupts intra-class compactness for tail classes, thereby degrading the overall model's performance (see Figure~\ref{fig:embedding_collapse}).

Building upon AM-Softmax, we introduce a specialized objective termed \emph{dual-margin penalization loss}. Our formulation explicitly rebalances the influence of head and tail classes through a dual-margin penalization mechanism. In doing so, we extend AM-Softmax to better accommodate fine-grained, long-tailed plant taxonomy. Since AM-Softmax promotes compact intra-class clustering and enhanced inter-class separability, these geometric properties further contribute to improved robustness under domain shifts and open-set conditions, as demonstrated in our experiments.

\vspace{-2mm}
\subsection{Dual-Margin Penalization Loss}
We propose a dual-margin penalization loss that operates at the logit level, applying margin constraints to both target and non-target classes as follows:
\begin{equation}
\mathcal{L}_{\mathrm{ours}} =
- \log
\frac{
    e^{s (z_y - m_y)}
}{
    e^{s (z_y - m_y)}
    + \sum_{k \ne y} e^{s (z_k - m_k)}
}.
\label{eq:cr2m}
\end{equation}
\noindent
where \( m_y = m_0 + \Delta_y \) is the \emph{target margin} for the ground-truth class \( y \), and \( m_k = \Delta_k \) is the \emph{non-target margin} for competing classes \( k \ne y \). The term \( m_0 \) is a pre-configured base margin. Here, \( \Delta_k \) denotes a learnable class-relative margin parameter jointly optimized during training. 
Specifically, \( \Delta_k \) is defined as
\begin{equation}
\Delta_k = m_0 \cdot \left( \frac{\rho_k}{m_0} \right)^{\zeta(\gamma)},
\label{eq:delta}
\end{equation}
\noindent where \( \gamma \) is a learnable parameter regularized via
\begin{equation}
\mathcal{L}_{\text{reg}} = \sum_{k \in \mathcal{C}} \left(\Delta_k - \rho_k\right)^2.
\label{eq:reg}
\end{equation}
$\rho_k$ is the log-normalized class statistic defined by $\rho_k \propto m_0 \cdot \left( -  \log \frac{N_k}{N} \right)$ such that $\sum_k \rho_k = m_0$, where $N_k$ is the number of samples in class \( k \) and \( N \) is the total number of samples across the class set \( \mathcal{C} = \{1,\dots,c\} \). Accordingly, \( \Delta_k \in [0,m_0] \) increases as \( N_k \) decreases, assigning larger margin adjustments to rare classes. The function \( \zeta(\cdot) \) is a smooth, monotonic transformation; in our experiments, we adopt the soft-plus parameterization \( \zeta(\gamma)=\log(1+e^\gamma) \). 

In principle, when a tail-class example is sampled as the target, a larger $m_y$ strengthens the attractive gradient, pulling its embeddings closer to the corresponding prototype and thereby improving intra-class compactness. On the other hand, since head-class samples dominate the training batches, tail classes frequently serve as non-targets. In this case, the non-target margin $m_k = \Delta_k$, which is larger for tail classes, reduces the effective logit $z_k - m_k$ in the softmax denominator, suppressing the repulsive gradient that would otherwise push the tail-class prototype away from its class mean. Without such regulation, the cumulative repulsive forces from abundant head-class samples may persistently shift tail-class prototypes, destabilizing optimization. We formally analyze this effect in Section~\ref{sec:dmloss}.

\vspace{-2mm}
\paragraph{Training Objective.}
The overall training objective is defined as
\begin{equation}
\mathcal{L} = \mathcal{L}_{\mathrm{ours}} + \lambda \mathcal{L}_{\mathrm{reg}},
\end{equation}
where \( \lambda \) controls the strength of regularization. In practice, training TaxoNet requires tuning only the base margin \( m \), which governs embedding compactness and separability, while the class-relative margin \( \Delta_k \) is a learnable parameter based on empirical class priors. The scaling factor \( s \) is fixed to 32 as large values of \( s \) may exacerbate gradient imbalance under long-tailed distributions, potentially hindering convergence for rare classes. The training algorithm and parameter configurations are provided in the Appendix.

\vspace{-2mm}
\subsection{Norm-Guided Sample Selection}
\label{sec:norm-guided}
\noindent Naïve oversampling of extremely scarce tail-class examples often induces memorization, leading to overfitting. Instead, we prioritize informative samples (i.e., those with low embedding norm magnitude), drawing inspiration from margin-based face representation learning \cite{meng2021magface, low2023slackedface}. Low-norm samples primarily arise from two scenarios: (1) underrepresented tail classes, and (2) samples exhibiting high intra-class variation due to seasonal or phenotypic changes (e.g., canopy structure, leaf morphology, or fruiting status), which may also occur within head classes. By emphasizing such samples, the model focuses on harder and more diverse instances, thereby improving generalization across ecological conditions while mitigating memorization from repeated tail-class oversampling. Importantly, this sampling strategy is adaptive: as embeddings evolve during training, previously emphasized samples tend to increase in norm magnitude and thus become less likely to be reselected under the norm-guided criterion.

\vspace{-2mm}
\paragraph{Oversampling with Augmented Diversity.} 
To enhance tail-class representation, we augment each training batch of \( B \) images by sampling an additional \( b \) tail-class instances using \textbf{AugMix}~\cite{hendrycks2020augmix}, applied stochastically with Bernoulli probability \( p \). From the \( B + b \) candidates, we retain the \( B \) samples with the lowest embedding norms. This norm-guided criterion is particularly compatible with margin-based softmax formulations, including our proposed dual-margin loss, where embedding geometry correlates with classification confidence: lower norms indicates less confident predictions due to weaker alignment with class prototypes. However, standard softmax loss does not impose such geometric calibration during optimization ~\cite{wang2018cosface,meng2021magface}. We show in Section~\ref{sec:experiments} that low-norm samples are strongly associated with tail classes or instances exhibiting substantial intra-class variability.

\section{Theoretical Analysis}
\label{sec:dmloss}
We establish theoretical grounds showing that the proposed dual-margin loss mitigates prototype deviation from class-conditional means by characterizing the mean-seeking behavior of class prototypes and analyzing how head-class dominance causes misalignment of tail classes.

\vspace{-2mm}
\paragraph{Definitions}
Our task is to divide a class index set $\mathcal{C}$ by $\mathcal{C} = \mathcal{C}_H \cup \mathcal{C}_T$ where $\mathcal{C}_{H}$ is the index set for head classes and $\mathcal{C}_T$ is for tail classes. The requirement only concerns that both index sets are non-empty, and the sample size of each head class is greater than that of any tail class. Accordingly, we divide the sample index set to $\{1,\dots, N\} = I_H \cup I_T$ where $I_H$ corresponds to head class samples, and $I_T$ to tail class samples. For notational simplicity, we let $\mathcal{L} = \mathcal{L}_{ours}$.

\vspace{-2mm}
\subsection{Prototype Alignment}
Under the update rules based on stochastic gradient descent (SGD), each class prototype $\hat{\mathbf{w}}_c$ tends to be in the direction of the gradient of the loss $\mathcal{L}$
\begin{equation}
\label{eq:prototype_update}
- \frac{\partial \mathcal{L}}{\partial \hat{\mathbf{w}}_c}
= -  
\sum_{i \not\in I_c} 
\frac{\partial \mathcal{L}_i}{\partial \hat{\mathbf{w}}_c}
- 
\sum_{i \in I_c}
\frac{\partial \mathcal{L}_i}{\partial \hat{\mathbf{w}}_c},
\end{equation}
where $\mathcal{L}_i$ is the loss for $i$-th sample, and $I_c$ is the set of indices of $c$-th class samples. The first term on the right-hand side (RHS) of Eq.~\eqref{eq:prototype_update} contributes to the undesired deviation of the prototype $\hat{\mathbf{w}}_c$.
When $\hat{\mathbf{w}}_c$ is a prototype of the tail class, its gradient is dominated by the first term as there are a much greater number of samples that are not of the $c$-th class. Without being properly handled, this causes unstable training and, therefore, poor representation. This will be discussed in the next part. 
The second term of the RHS of Eq.~\eqref{eq:prototype_update}, on the other hand, aligns the prototype $\hat{\mathbf{w}}_c$ to be in the direction of the class-wise embedding mean $\boldsymbol{\mu}_c$ based on the following:
\begin{prop}
Let $I_c$ be the set of indices of samples in the $c$-th class. Then,
\begin{equation}
\lVert
\sum_{i \in I_c} \frac{\partial \mathcal{L}_i}{\partial \hat{\mathbf{w}}_c}
- \alpha \boldsymbol{\mu}_c
\rVert
\to 0
\end{equation}
as $\sigma_c \to 0$
where $\boldsymbol{\mu}_c$ is class-wise mean 
\begin{equation}
\boldsymbol{\mu}_c = \frac{1}{N_c} \sum_{i \in I_c} \hat{\mathbf{x}}_i,
\end{equation}
$\sigma_c$ is the class-wise standard deviation of the predicted probabilities
\begin{equation}
\sigma_c^2
= \frac{1}{N_c - 1}
\sum_{i \in I_c}
|
p_{i,c} - \overline{p}_c
|^2
\end{equation}
with $\overline{p}_c = \frac{1}{N_c} \sum_{i \in I_c} p_{i, c}$,
and $\alpha = N_c (1 - \overline{p}_c)$
\end{prop}

\begin{proof}
As
\[
\frac{\partial \mathcal{L}_i}{ \partial \hat{\mathbf{w}}_c} = (1 - p_{i, c}) \mathbf{x}_i,
\]
we examine the norm
\[
\left\lVert
\sum_{i \in I_c} \frac{\partial \mathcal{L}_i}{\partial \hat{\mathbf{w}}_c}
- \alpha \boldsymbol{\mu}_c
\right\rVert
= 
\left\lVert
\sum_{i \in I_c} 
(\overline{p}_c - p_{i,c}) \hat{\mathbf{x}}_i
\right\rVert.
\]
By the triangle inequality:
\begin{align}
\left\lVert
\sum_{i \in I_c} 
(\overline{p}_c - p_{i,c}) \hat{\mathbf{x}}_i
\right\rVert
&\leq 
\sum_{i \in I_c}
\left| \overline{p}_c - p_{i,c} \right|
\left\lVert \hat{\mathbf{x}}_i \right\rVert \nonumber \\
&\leq 
N_c \cdot \max_i \left| \overline{p}_c - p_{i, c} \right|,
\end{align}
which converges to 0 as the intra-class variance $\sigma_c \to 0$.
\end{proof}

\vspace{-2mm}
\subsection{Tail-Class Prototype vs Class-Wise Mean}
As indicated in the gradient decomposition of Eq.~\eqref{eq:prototype_update}, when $c \in \mathcal{C}_T$ is of a tail class, its prototype $\hat{\mathbf{w}}_c$ deviates from the class-wise mean $\boldsymbol{\mu}_c$ as a result of frequent updates from head class samples, through minimizing $\mathcal{L}_i$ with $i \in I_H$. In particular, the tail class prototype $\hat{\mathbf{w}}_c$ deviates from the class-wise mean $\boldsymbol{\mu}_c$ due to the head-class gradients $-\frac{\partial \mathcal{L}_i}{\partial \hat{\mathbf{w}}_c}$ for $i \in I_H$. The impact of this undesired deviation on the tail-class prototype is determined by the magnitude of the gradient, that is, $\lVert \frac{\partial \mathcal{L}_i}{\partial \hat{\mathbf{w}}_c} \rVert$. To learn properly on the tail classes, this impact must be constrained.

Here, we demonstrate that under the proposed dual-margin penalization loss, its influence is bounded by an exponential term that decays with the class margin $m_c$, leading to the following proposition:
\begin{prop}
\label{prop:prevent_dev}
For head-class loss $\mathcal{L}_i$ with $i \in I_H$ and tail-class prototype $\hat{\mathbf{w}}_c$ with $c \in \mathcal{C}_T$, we have
\begin{equation}
\lVert
\frac{\partial \mathcal{L}_i}{\partial \hat{\mathbf{w}}_c}
\rVert
\leq
\exp( m_{y_i} - m_c)
\end{equation}
if $y_i = \arg\max_k p_{i,k}$.
\end{prop}
This condition in the proposition holds for most samples, even during early training, as minimizing loss $\mathcal{L}_i$ inherently maximizes $p_{i, y_i}$. If not for simplicity, the proposition can be proved in a much weaker assumption than $\arg\max_k p_{i,k} \in \mathcal{C}_H \setminus \{ c \}$.
\begin{proof}
Observe
\begin{equation}
p_{i,c}
\leq
\frac{\exp(z_{i, c} - m_c)}{\exp(z_{i, y_i}- m_{y_i}) }
\leq \exp(m_{y_i} - m_c)
\end{equation}
this holds because the sample loss $\mathcal{L}_i$ corresponds to the head classes (i.e., $i \in I_H$). Consequently, the gradients become $- \partial \mathcal{L}_i / \partial \hat{\mathbf{w}}_c = p_{i, c} \hat{\mathbf{x}}_i$ with $\lVert \hat{\mathbf{x}}_i \rVert = 1$.
\end{proof}
\noindent
Therefore, for tail classes, the prototype is less influenced by head-class learning.


\section{Evaluation}
\label{sec:experiments}

\vspace{-2mm}
\subsection{Experimental Setup}
\label{sec:setup}

\paragraph{Datasets.} We evaluate on three plant datasets: Google Auto-Arborist (AA)~\cite{beery2022auto}, iNat-Plantae (iNaturalist-2019 restricted to the only Plantae kingdom)~\cite{van2018inaturalist}, and NAFlora-Mini~\cite{park2024naflora}. AA contains urban street-tree images from three North American regions: AA-West, AA-Central, and AA-East; iNat-Plantae includes citizen-science plant observations across global ecosystems; and NAFlora-Mini comprises herbarium specimens of North American vascular plants. Dataset statistics are provided in the Appendix. 

\vspace{-2mm}
\paragraph{Implementation Details.} We employ the ResNet-101 backbone pretrained on ImageNet as the embedding encoder. For dual-margin penalization, we set the scaling factor to \( s = 32.0 \) and the base margin to \( m = 0.15 \). All input images are resized to \( 614 \times 512 \times 3 \). The batch size is set to \( B = 32 \), oversampled with \( b = 8 \) additional tail-class examples using \textbf{AugMix}~\cite{hendrycks2020augmix}. We set the Bernoulli probability to \(p=0.1\) for stochastic oversampling.

\vspace{-2mm}
\paragraph{Comparison and Evaluation Metrics.}
We compare our model with the standard cross-entropy (CE) classifier, the margin-based AM-Softmax~\cite{wang2017normface}, and established state-of the-art (SOTA) models: class-balanced loss (CBL)~\cite{cui2019class}, logit adjustment (LA)~\cite{menon2020long}, and label-distribution-aware margin loss (LDAM)~\cite{cao2019learning}. We also benchmark against multimodal large language models (MLLMs) such as \textsc{GPT-4o}~\cite{hurst2024gpt} and \textsc{Gemini-2.5-Flash}~\cite{team2023gemini}, as well as vision–language foundation model (VLFM), \textsc{BioCLIP}~\cite{stevens2024bioclip}, for zero-shot plant classification.

We report rank-1 accuracy (overall performance) and macro-recall (average per-class accuracy) to capture performance across both head and tail taxa. For open-set recognition, we use true negative rate (TNR) and overall accuracy (ACC), both evaluated at a fixed true positive rate (TPR) of 95\%.

\subsection{Results}
\paragraph{Performance Comparison.} Table~\ref{tab:main_experimental_results} summarizes that TaxoNet achieves significant macro-averaged recall gains, outperforming LDAM (the strongest SOTA baseline) by 6\% on Google AA, 3\% on iNat-Plantae, and 1\% on NAFlora-Mini. Although rank-1 accuracy shows marginal decline, this trade-off reflects improved balance between head and tail classes. Built on AM-Softmax~\cite{wang2018cosface}, our model explicitly captures fine-grained class geometry in the embedding space. Meanwhile, LDAM, LA, and CBL improve tail-class decision boundaries through margin penalties, logit adjustments, and reweighting strategies, but they lack mechanisms for enforcing within-class compactness and between-class separation, often resulting in embedding drift. We examine other open-world challenges next.

\begin{table*}[t]
\caption{Performance comparison across benchmarking datasets in terms of rank-1 accuracy (R@1) and macro-averaged recall (\%), including ablations of TaxoNet under varying configurations. For iNat-Plantae, R@1 is effectively equivalent to recall as the number of test sample is similar across classes. The best and second-best results are highlighted in \textbf{bold} and \underline{underlined}, respectively, across all tables in this paper.}
\label{tab:main_experimental_results}
\centering  
\resizebox{0.99\textwidth}{!}{ 
\renewcommand{\arraystretch}{1.2}
\begin{tabular}{p{3.5cm} 
>{\centering\arraybackslash}p{1.2cm}  
>{\centering\arraybackslash}p{1.2cm}  
>{\centering\arraybackslash}p{1.2cm}  
>{\centering\arraybackslash}p{1.2cm}
>{\centering\arraybackslash}p{1.2cm}  
>{\centering\arraybackslash}p{1.2cm}
cc c cc}
\toprule
\multirow{2}{*}{\textbf{Models}} 
& \multicolumn{2}{c}{\textbf{AA-Central}}
& \multicolumn{2}{c}{\textbf{AA-West}}
& \multicolumn{2}{c}{\textbf{AA-East}} 
& \multicolumn{1}{c}{\multirow{1}{*}{\textbf{iNat-Plantae}}} 
& \multicolumn{2}{c}{\multirow{1}{*}{\textbf{NAFlora-Mini}}} \\
& R@1 & Recall & R@1 & Recall & R@1 & Recall
& R@1$=$Recall & R@1 & Recall \\   
\midrule   
Standard CE & 91.59 & 63.55 & 83.44 & 59.24 & 82.87 & 56.21 & 79.28 & 89.48  & 86.29 \\ 
CBL~\cite{cui2019class} & 91.46 & 67.59 & 84.36 & 62.72 & 84.12 & 63.44 & 81.38 & 90.88 & 90.08 \\
LDAM~\cite{cao2019learning} & \textbf{92.31} & 67.85 & \textbf{85.96}  & 62.82 & \textbf{85.32} & 62.21 & 81.57 & 91.26 & 89.96 \\   
LA~\cite{menon2020long} & 90.30 & 66.90 & 84.08 & 61.95 & 83.32 & 61.21 & 80.30 & 90.73 & 89.98 \\
AM-Softmax \cite{wang2018cosface} & 92.07 & 64.77 & 84.99 & 60.88 &  83.33 & 59.22 & 80.25 & 89.84 & 88.85  \\  
\midrule 
TaxoNet w/ A, C & \underline{92.12} & 63.91 & 84.80 & 60.81 & 83.77 & 60.72  & 78.91 & 89.84 & 88.85  \\
TaxoNet w/ B, C & 91.30 & 67.59 &  84.37 & 63.15  & 83.17 & 62.25 & 80.01 & 90.55 & 89.56\\
TaxoNet w/ B, D & 92.00 & 67.56 & 84.36 & 63.12 & 84.42 & 62.47 & 81.08 & 91.10 & 89.83 \\  
TaxoNet w/ B, E & 91.84 & \underline{69.48} & 85.72 & \underline{65.49} & \underline{84.53} & \underline{63.46} & \underline{81.87} & \underline{91.48} & \underline{90.16} \\
\textbf{TaxoNet w/ B, E, F} & 91.92 & \textbf{72.90} & \underline{85.94} & \textbf{67.67} & 84.32 & \textbf{64.96} & \textbf{83.21} & \textbf{91.52} & \textbf{90.40} \\
\bottomrule   
\end{tabular}      
} 
\smallskip\\
\parbox{\linewidth}{\small 
\textbf{Note.} A: Base Margin Only; B: Dual-Margin; C: No Oversampling (OS); D: OS with Random Selection; E: OS with Norm-Guided Selection; F: Regularization.}
\end{table*}

\paragraph{Domain Generalization.}
We examine regional distribution shift by transferring the AA-Central models of Auto-Arborist to other regions for our model, CBL, and LDAM. Due to taxonomic divergence, performance is evaluated using instance-level matching. As shown in Table~\ref{tab:domain_generalization}, our model demonstrates stronger domain generalization, attributed to norm-guided selection, whereas CBL and LDAM rely on random selection.

\begin{table}[h]
\caption{Rank-1 accuracy and macro-recall (\%) comparison evaluating regional distribution shift for the model trained on \textbf{AA-Central}.}
\label{tab:domain_generalization}
\centering
\scriptsize
\resizebox{0.7\columnwidth}{!}{
\renewcommand{\arraystretch}{1.2}
\begin{tabular}
{
p{1.5cm}
>{\centering\arraybackslash}p{1.2cm}  
>{\centering\arraybackslash}p{1.2cm}  
>{\centering\arraybackslash}p{1.2cm}  
>{\centering\arraybackslash}p{1.2cm}  
>{\centering\arraybackslash}p{1.2cm}
>{\centering\arraybackslash}p{1.2cm} 
}
\toprule
& \multicolumn{2}{c}{\textbf{AA-Central}} & \multicolumn{2}{c}{\textbf{AA-West}} & \multicolumn{2}{c}{\textbf{AA-East}} \\
& R@1 & Recall & R@1 & Recall & R@1 & Recall \\ 
\midrule     
CBL  & \underline{91.18} & 65.93  & 62.73 & 44.15 & 47.19 & 37.71 \\ 
LDAM & \textbf{91.41} & \underline{68.37} & \textbf{65.95} & \underline{46.50} & \underline{49.26} & \underline{39.95} \\
TaxoNet  & 91.14 & \textbf{70.18} & \underline{64.43} & \textbf{48.10} & \textbf{52.13} & \textbf{40.53} \\
\bottomrule      
\end{tabular} 
}  
\end{table}

\vspace{-2mm}
\paragraph{Open-Set Recognition.}
In open-set evaluation, the goal is to flag previously unseen taxa for expert review. Taxonomists assess whether the sample belongs to a known class or represents a novel taxon. Our results in Table~\ref{tab:openset} shows that our model achieves over 90\% TNR on Regions West and Central of Auto-Arborist, outperforming CBL and LDAM in unknown-class rejection. This comes from its base margin, which improves embedding separation. Increasing the margin can further improve unknown detection, though it may marginally reduce performance on known classes.

\begin{table}[h]
\caption{Performance comparison in terms of TPR, TNR, and acc. for open-set evaluation on \textbf{AA-Central}, with the threshold calibrated to 95\% TPR. An unknown set (88 novel taxa, 8,845 samples) is used to simulate open-set conditions.}
\label{tab:openset}
\centering
\resizebox{0.6\columnwidth}{!}{
\renewcommand{\arraystretch}{1.2}
\begin{tabular}
{
p{1.5cm}
>{\centering\arraybackslash}p{1.2cm}  
>{\centering\arraybackslash}p{1.2cm}  
>{\centering\arraybackslash}p{1.2cm}  
>{\centering\arraybackslash}p{1.2cm}  
>{\centering\arraybackslash}p{1.2cm}
>{\centering\arraybackslash}p{1.2cm} 
}
\toprule
& \multicolumn{3}{c}{\textbf{AA-Central}} & \multicolumn{3}{c}{\textbf{AA-West}} \\
& TPR & TNR & Acc. & TPR & TNR & Acc. \\ 
\midrule       
CBL  & 91.83 & 86.34 & 89.08 &  86.93 & 84.76 & 85.84  \\ 
LDAM  & 92.29 & 88.59 & \underline{90.44} & 88.84 & 87.09 & \underline{87.96} \\ 
TaxoNet  & 93.25 & 91.28 & \textbf{92.26} & 90.91 & 89.84 & \textbf{90.38} \\
\bottomrule           
\end{tabular} 
} 
\end{table} %

\begin{table}[h]
\caption{Performance comparison using macro-averaged recall (\%) across taxonomic levels for TaxoNet, CE, CBL, and LDAM on \textbf{AA-Central}. \textbf{Taxa group} is defined by training cardinality: head ($>$2,000 samples), between (100–2,000 samples), and tail ($<$100 samples). Class-level recall results are reported in the Appendix.}
\label{tab:hierachical}
\centering
\resizebox{0.6\columnwidth}{!}{
\renewcommand{\arraystretch}{1.2}
\begin{tabular}{
>{\centering\arraybackslash}p{2.5cm}  
>{\centering\arraybackslash}p{2.2cm}  
>{\centering\arraybackslash}p{1.2cm}  
>{\centering\arraybackslash}p{1.2cm}  
>{\centering\arraybackslash}p{1.2cm}  
>{\centering\arraybackslash}p{1.2cm}  
}
\toprule
 & \textbf{Taxa Group} & 
\textbf{TaxoNet} & \textbf{CE} &\textbf{CBL} & \textbf{LDAM} \\
\midrule
\multirow{1}{*}{\textbf{Family}} 
& \makecell{Overall}  & \textbf{78.70} & 72.67 & \underline{74.57} & 72.09 \\
\midrule
\multirow{4}{*}{\textbf{Genus}} 
& Head     & 92.37 & 92.60 & \underline{93.34} & \textbf{94.13} \\
& Between  & \underline{86.13} & 81.60 & 84.80 & \textbf{86.80} \\
& Tail     & \textbf{57.06} & 43.18 & \underline{50.27} & 44.18 \\
\cline{2-6}
& Overall  & \textbf{72.90} & 63.55 & \underline{67.59} & 66.85 \\
\bottomrule
\end{tabular}
}
\end{table}

\paragraph{Additional Insights.} We compare the performance of our model with CE, CBL~\cite{cui2019class}, and LDAM~\cite{cao2019learning} at both family and genus levels in Table~\ref{tab:hierachical}. Our model marks the top performance, especially in tail groups, while LDAM outperforms in head classes. This reflects the effect of norm-guided sampling, which prioritizes tail examples, unlike random sampling in LDAM. Interestingly, our embeddings reflect taxonomic relationships for hierarchical classification—a behavior that we leave for future exploration.

\vspace{-2mm}
\subsection{Analysis}
\paragraph{Ablation Studies.} 
We analyze the contribution of each model component: base margin, class-relative margin (with and without power-based regularization), and norm-guided sample selection. As shown in Table~\ref{tab:main_experimental_results}, their combination yields the highest gain. Notably, power-based regularization stabilizes the base margin by adapting class-relative margins to prior statistics and learned signals, improving generalization in class imbalanced and fine-grained settings.

\begin{figure}[h]
    \centering
    \begin{minipage}{0.5\linewidth}
        \centering
        \includegraphics[width=\linewidth]{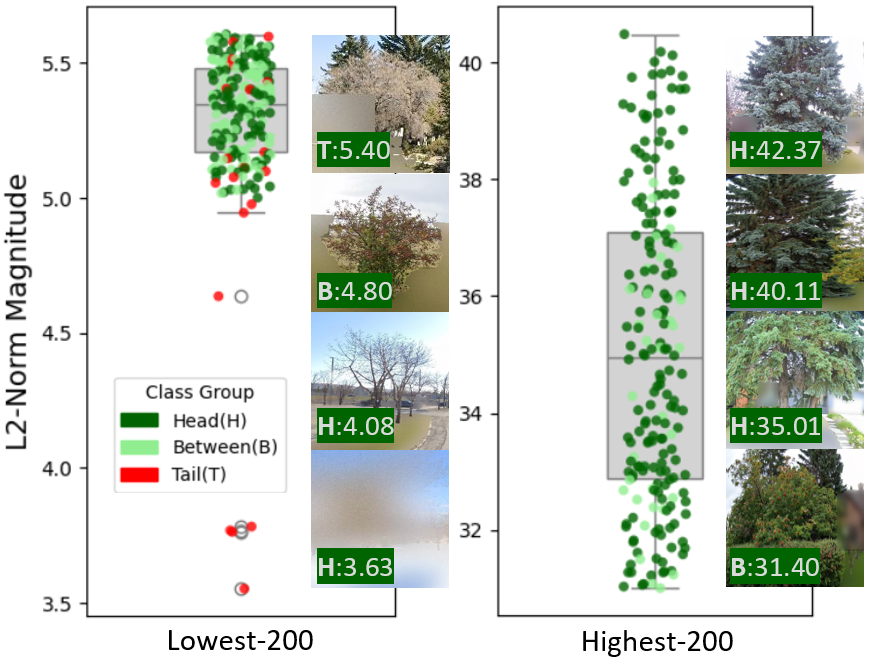}
    \end{minipage}
    \hfill
    \begin{minipage}{0.45\linewidth}
        \captionsetup{type=figure}
        \caption{Embedding norm distribution for the 200 highest-norm and 200 lowest-norm samples in \textbf{AA-Central}. High-norm examples are concentrated in head and between classes; low-norms are from tail or those with greater within-class variations (e.g., due to seasonal changes). \textbf{Class Group} is defined by training cardinality: head $>$ 2,000 samples, tail $<$ 100 samples, between otherwise.}
        \label{fig:norm_analysis}
    \end{minipage}
\end{figure}

Figure~\ref{fig:norm_analysis} shows that high-norm instances primarily originate from head classes or confidently learned examples with distinct taxonomic traits. Conversely, low-norm instances typically arise from tail classes or samples with high within-class variation due to seasonal and morphological changes. As reported in Table~\ref{tab:main_experimental_results}, these findings validate our norm-guided selection strategy, which prioritizes low-norm samples to enhance tail-class generalization (Table~\ref{tab:hierachical}) and mitigate overfitting from naïve oversampling. TaxoNet with dual-margin and norm-guided selection consistently outperforms random selection. We reemphasize that \emph{the norm proxy applies only to models built on margin-based softmax variants~\cite{meng2021magface}, including our dual-margin penalization loss, and is not compatible with baselines such as CBL, LDAM, or LA.}

\begin{figure}[t]
    \centering
    \begin{minipage}{0.58\linewidth}
        \centering
        \includegraphics[width=\linewidth]{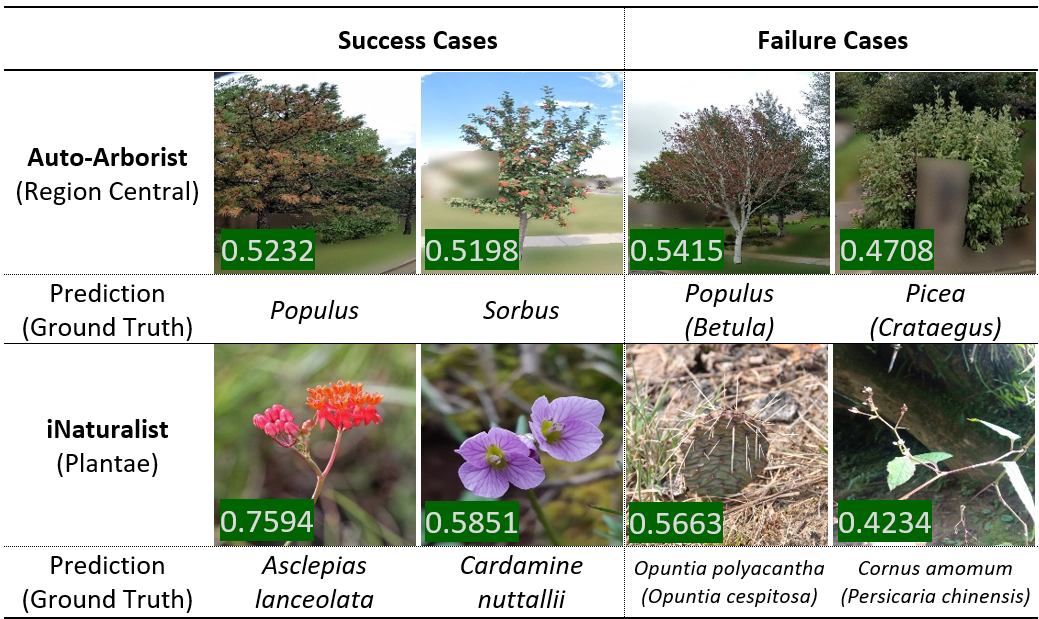}
    \end{minipage}
    \hfill
    \begin{minipage}{0.38\linewidth}
        \captionsetup{type=figure}
        \caption{Success and failure cases of model predictions on AA-Central and iNat-Plantae. Softmax prediction scores are shown for each test image.}
        \label{fig:success_failure}
    \end{minipage}
\end{figure}

\vspace{-2mm}
\paragraph{Success and Failure Cases.}
We analyze success and failure cases in Figure \ref{fig:success_failure}. Our model performs well under challenging conditions such as seasonal variation in morphology and flower-only inputs that lack contextual cues like leaves or growth form. Failure cases often involve visually ambiguous samples: \emph{Opuntia polyacantha} is misclassified as \emph{Opuntia cespitosa}, a closely related species from the same genus, and \emph{Cornus amomum} as \emph{Persicaria chinensis}, likely due to leaf-only views. These limitations underscore the need for future multi-view, multimodal models, potentially leveraging advances in MLLMs or VLFMs, to support reasoning.

\begin{table}[t]
\caption{
\textbf{Comparative analysis of taxonomic classification performance.} We report macro-averaged recall (\%) on the iNat-Plantae dataset, where {taxa group} is defined by training cardinality:  10 head taxa ($>$500 samples), 10 between taxa (50--500 samples), and 20 tail taxa ($<$50 samples). TaxoNet consistently outperforms both general-purpose MLLMs and domain-specific vision-language models, particularly in the rare-taxa (tail) regime. See Appendix for the class-level results.}
\label{tab:llm_inaturalist}
\centering
\resizebox{0.8\columnwidth}{!}{
\renewcommand{\arraystretch}{1.2}
\begin{tabular}{
>{\centering\arraybackslash}p{2.5cm}
>{\centering\arraybackslash}p{2.2cm} 
>{\centering\arraybackslash}p{2cm}
>{\centering\arraybackslash}p{1.2cm}
>{\centering\arraybackslash}p{1.2cm}
>{\centering\arraybackslash}p{1.2cm}
>{\centering\arraybackslash}p{1.2cm}
>{\centering\arraybackslash}p{1.4cm}}
\toprule
\multirow{2}{*}{\makecell{}} & 
\multirow{2}{*}{\makecell{\textbf{Taxa Group}}} &
\multirow{2}{*}{\textbf{TaxoNet}} &
\multicolumn{2}{c}{\textbf{GPT-4o}} & 
\multicolumn{2}{c}{\textbf{Gemini-2.5}} &
\textbf{BioCLIP} \\
\cline{4-8}
& & & \textbf{ZS} & \textbf{CoT} & \textbf{ZS} & \textbf{CoT} & \textbf{ZS} \\
\toprule
\multirow{4}{*}{\textbf{Genus}} 
& Head(10)     & \textbf{96.53} & 92.16 & 91.20 & 91.67 & \underline{92.46} & 60.34 \\ 
& Between(10)  & \textbf{94.91} & 87.23 & 88.62 & 88.75 & \underline{88.91} & 48.00 \\ 
& Tail(20)     & \textbf{96.71} & 91.68 & 91.87 & 92.60 & \underline{93.11} & 51.92 \\ 
\cline{2-8}
& Overall      & \textbf{95.96} & 89.08 & 89.91 & 90.86 & \underline{91.50} & 53.42 \\ 
\midrule
\multirow{4}{*}{\textbf{Species}} 
& Head(10)     & \textbf{96.67} & 13.00 & 13.00 & 13.33 & 16.67 & \underline{59.26} \\ 
& Between(10)   & \textbf{93.33} & 43.33 & 50.00 & 46.67 & \underline{53.34} & \underline{53.34} \\ 
& Tail(20)     & \textbf{76.67} & 16.67 & 11.67 & 26.67 & 31.67 & \underline{48.33}\\ 
\cline{2-8}
& Overall      & \textbf{83.21} & 35.28 & 32.51 & 40.39 & 41.46 & \underline{52.14} \\ 
\bottomrule
\end{tabular}}
\end{table}

\vspace{-2mm}
\subsection{Further Benchmarking}
\paragraph{Comparison with MLLMs.} 
Given general-purpose Multimodal Large Language Models (MLLMs) excel at broad tasks \cite{team2023gemini, hurst2024gpt}, we test whether they can handle biodiversity recognition. Table~\ref{tab:llm_inaturalist} compares our model against GPT-4o~\cite{hurst2024gpt} and Gemini-2.5~\cite{team2023gemini} on the iNat-Plantae dataset. To evaluate these models, we used zero-shot (ZS) persona-based prompts and asked the AI to act as a botanical expert \cite{shanahan2023role} and a chain-of-thought (CoT) approach reasoning \cite{wei2022chain}. This guided the models to reason through the taxonomic hierarchy, first identifying the genus before narrowing down to the specific species.

TaxoNet significantly outperforms MLLMs across both genus and species levels, achieving 2$\times$ higher performance in species-level accuracy. Although CoT reasoning improves over ZS prompting in Gemini's results, it yields negligible gains for GPT-4o, highlighting the inherent limitations of general-purpose models in fine-grained botanical discrimination. We provide additional results for AA-Central and prompting templates in the Appendix.

\vspace{-2mm}
\paragraph{Comparison with VLFM.}
Unlike open-domain MLLMs, \textsc{BioCLIP} leverages large-scale biodiversity resources spanning plants, animals, fungi, and microbes, and is therefore not strictly domain-specialized. 
As \textsc{BioCLIP} includes training examples overlapping with iNat-Plantae, it attains higher species-level performance than MLLMs. However, MLLMs, aided by text-guided CoT reasoning, enhances interpretability at coarser taxonomic levels. Overall, our proposed domain-specialized model establishes a strong foundation for open-world taxonomic monitoring when deployed in the plant domain.

\section{Conclusion}
We propose a dual-margin penalization objective that balances learning signals across dominant and rare taxa, forming the basis of \emph{TaxoNet}, an embedding framework for plant taxonomy recognition. Validated across urban, citizen-science, and herbarium plant datasets, TaxoNet consistently outperforms baselines, including state-of-the-art multimodal foundation models.

TaxoNet represents a significant advancement in fine-grained and long-tailed representation learning tailored to the structural complexities of ecological data. By addressing additional open-world deployment challenges such as spatiotemporal shifts, hierarchical taxonomic structure and novel taxa, our framework enables scalable monitoring of plant diversity across urban and natural forest ecosystems. Furthermore, it provides a robust backbone for citizen science platforms such as \emph{iNaturalist}, \emph{Flora Incognita}, \emph{Pl@ntNet}). 
We anticipate that our model can be extended to broader biodiversity monitoring tasks.


\bibliographystyle{splncs04}
\bibliography{sections/7_references}



\end{document}